%% file: main.tex
\documentclass[conference]{IEEEtran}
\IEEEoverridecommandlockouts

\usepackage{cite}
\usepackage{amsmath,amssymb,amsfonts}
\usepackage{algorithmic}
\usepackage{graphicx}
\usepackage{textcomp}
\usepackage{xcolor}
\usepackage{subcaption}

\usepackage[hidelinks]{hyperref}

\def\BibTeX{{\rm B\kern-.05em{\sc i\kern-.025em b}\kern-.08em
    T\kern-.1667em\lower.7ex\hbox{E}\kern-.125emX}}
\begin{document}

\title{Can a Vision-Language Model Tune Clustering In Context?}

\author{\IEEEauthorblockN{
Javith Ranatunga\textsuperscript{*}}
\IEEEauthorblockA{
\textit{Riverside STEM Academy}\\
Riverside, CA, USA \\
javithranatunga@gmail.com}
\and
\IEEEauthorblockN{
Paimon Goulart}
\IEEEauthorblockA{
\textit{University of California, Riverside}\\
Riverside, CA, USA \\
pgoul002@ucr.edu}
\and
\IEEEauthorblockN{
Evangelos E. Papalexakis}
\IEEEauthorblockA{
\textit{University of California, Riverside}\\
Riverside, CA, USA \\
epapalex@cs.ucr.edu}
}

\maketitle

\begingroup
\renewcommand{\thefootnote}{\fnsymbol{footnote}}
\footnotetext[1]{High school student; work completed as part of high school final capstone project.}
\endgroup

\begin{abstract}

Clustering is a widely used unsupervised learning method, but selecting an appropriate algorithm and set of hyperparameters for a specific dataset can require many possible configurations and evaluations. In this work, we investigate whether a Vision-Language Model can instead visually evaluate clustering results and use those decisions to then tune an explicit clustering algorithm. We study this through three progressive experiments using Qwen3.5-27B and five controlled synthetic clustering scenarios. First, we test whether the VLM can distinguish the best clustering result from a pair of plots. Second, we test whether it can assign numerical scores to those plots where a higher score represents a better result. Finally, we use these judgments within an agentic loop in which the VLM scores a k-means clustering and proposes updated hyperparameters that may result in better clustering. Our exploratory results show that the VLM performs well when clear cluster structure is present, while struggling on more ambiguous and noisy settings. In the full agentic experiment, the VLM produces clustering configurations with silhouette scores relatively close to those of grid search and Bayesian optimization, while terminating in much fewer iterations. Overall, these results provide an initial proof of concept that VLM-based visual feedback can be used to autonomously tune an explicit clustering algorithm.

\end{abstract}

\begin{IEEEkeywords}
clustering, vision-language models, in-context learning, hyperparameter optimization, agentic AI
\end{IEEEkeywords}

\input{sec/introduction}
\input{sec/related_works}
\input{sec/exploratory_study}
\input{sec/methods}
\input{sec/experiments}
\input{sec/conclusion}
\input{sec/acknowledgments}

\bibliography{BIB/refs}
\bibliographystyle{ieeetr}

\end{document}

%% file: sec/introduction.tex
\section{Introduction}

Clustering is a fundamental unsupervised learning method that is used as an exploratory technique to identify patterns within a typically unlabeled dataset \cite{JAIN2010651}. By grouping similar points of data, clustering can reveal meaningful subpopulations or outliers that are hard to detect at first, providing researchers with a better understanding of their data. We can then use these insights to further investigate said groups, which could reveal patterns that would have initially gone unnoticed. For these reasons, clustering has become common practice across many fields, proving crucial in many scientific discoveries. In biology, clustering is commonly used in cellular analysis to identify distinct cell types, populations, and states, including heterogeneous populations within tumors \cite{zhang2019probabilistic}. In physics, clustering is used to identify and categorize different types of noise/glitches within gravitational-wave detectors such as LIGO \cite{10825388}. In social networks, clustering is used in order to identify different communities, providing us with groups of users that are more connected with one another than the rest of the network \cite{CHUNAEV2020100286}.

Despite the value clustering brings across many domains, finding the right clustering method and configuration is often a difficult task. There are many different clustering methods to choose from, including k-means, DBSCAN, HDBSCAN, and spectral clustering. Each comprises its own hyperparameters, producing a large combinatorial space of different configurations to choose from. Clustering itself is also highly context-sensitive \cite{8215642}, a method that performs well on one dataset or domain may not produce meaningful clustering in another. As a result, searching over different algorithms and hyperparameters is often necessary. Evaluating these configurations introduces the new challenge of what constitutes a "good" clustering result. Popular methods, such as the elbow method, have been heavily scrutinized for producing ambiguous results. Other methods, such as the silhouette coefficient, may favor certain aspects of a clustering result, leading to conflicts with other evaluation metrics  \cite{Schubert_2023}. As a consequence, traditional hyperparameter optimization methods that aim to automate this process (e.g., Grid Search or Bayesian Optimization) require some metric or objective to optimize, even when unsure about which metric to use \cite{pmlr-v139-gosgens21a}. Therefore, manual review of the clustering structure is often integrated into the clustering pipeline to evaluate separation, overlap, and other intrinsic properties that ultimately dictate the selection of the primary clustering model.

Since researchers may need to manually inspect and visually verify clustering results through plots, is it possible to instead automate this process using a vision-language model (VLM)? Rather than relying solely on a predefined clustering metric or requiring researchers to repeatedly inspect each result, a VLM can visually analyze the clustering and directly propose changes to its hyperparameters. However, a general-purpose VLM may not inherently understand what constitutes a desirable clustering for a specific domain. Instead of fine-tuning the model on a large collection of domain-specific examples, we can use in-context learning (ICL) to provide examples of clustering results and their desired evaluations directly during inference time \cite{zhou-etal-2024-visual}. This allows for the VLM to become more domain adaptive without the need to retrain the model and, in principle, allow domain-specific examples to be exchanged when applying the framework to new datasets. Here, we focus on k-means as a proof of concept. Importantly, rather than directly giving the VLM raw data and directly produce clustering assignments, we use the VLM as an evaluator and controller that proposes changes to an explicit clustering algorithm. As a result, the clustering procedure remains interpretable as we can directly observe the changes the VLM made even if the VLM itself is a black box. More broadly, this provides an initial step towards using multimodal agents to adapt and potentially help design new clustering procedures while preserving the transparency of algorithm decisions, allowing for them to still be human interpretable.

In this work, we explore this idea through three progressive experiments. We test whether a VLM with in-context examples can (1) distinguish between a pair of clustering results, (2) numerically score clustering results, and (3) use these judgments to iteratively tune k-means hyperparameters within an agentic loop.

%% file: sec/related_works.tex
\section{Related Work}

Recent work has explored the capabilities of deep learning models for dataset clustering. Deep clustering methods use deep neural networks to learn representations that are better suited to identifying cluster structure, directly aiding clustering algorithms \cite{10585323}. More recently, there has been more of a growing push to incorporate foundation models within clustering. ClusterLLM uses LLM guidance to aid in text clustering \cite{zhang-etal-2023-clusterllm}, and IC\textbar TC uses foundation models to perform clustering based on specific instructions \cite{kwon2024ictc}. Notably, In-Context Clustering shows that foundation models can cluster numerical and image data in-context \cite{wang2025incontextclusteringlargelanguage}. These works all demonstrate the potential that foundation models can successfully aid in the task of clustering. However, due to the black-box nature of these foundation models, a big concern can arise when the foundation model itself is directly responsible for assigning clusters, mainly in interpretability. To mitigate this, we use a VLM to visually evaluate and propose changes to an explicit clustering algorithm. Facilitating the VLM to assist in clustering while still allowing the exact procedure to be directly observable.

More closely related, other recent work has used foundation models to evaluate and improve clustering results. Dial-In LLM uses an LLM to assess and improve intent clusters \cite{hong-etal-2025-dial}. LLMEdgeRefine and additional work use LLM reasoning to refine ambiguous text clusters \cite{feng-etal-2024-llmedgerefine, islam-2026-reasoning}. These methods all show that foundation models can be used to evaluate and improve on existing clustering results rather than directly perform clustering, but these approaches all operate over textual information rather than visually inspecting the plots of the clusters themselves.

Separately, recent work has explored using LLM agents for hyperparameter tuning and other more general ML tasks \cite{pmlr-v280-liu25c, AutoML_Agent}. In other closely related work, a VLM is used to provide feedback and insights from clustering plots \cite{inproceedings}. However, this feedback was not used to modify the clustering algorithms; rather, just given to researchers as analysis.

Distinctly from all the work stated above, our framework uses visual in-context examples to guide the VLM in visually evaluating clustering results and then iteratively proposing modifications to hyperparameters in an existing clustering algorithm. To our knowledge, this is the first work to use visual feedback from clustering plots in a closed loop to tune a clustering method.

%% file: sec/exploratory_study.tex
\section{Exploratory Study}

\subsection{Dataset}

\begin{figure*}[t]
    \centering
    \includegraphics[width=0.95\textwidth]{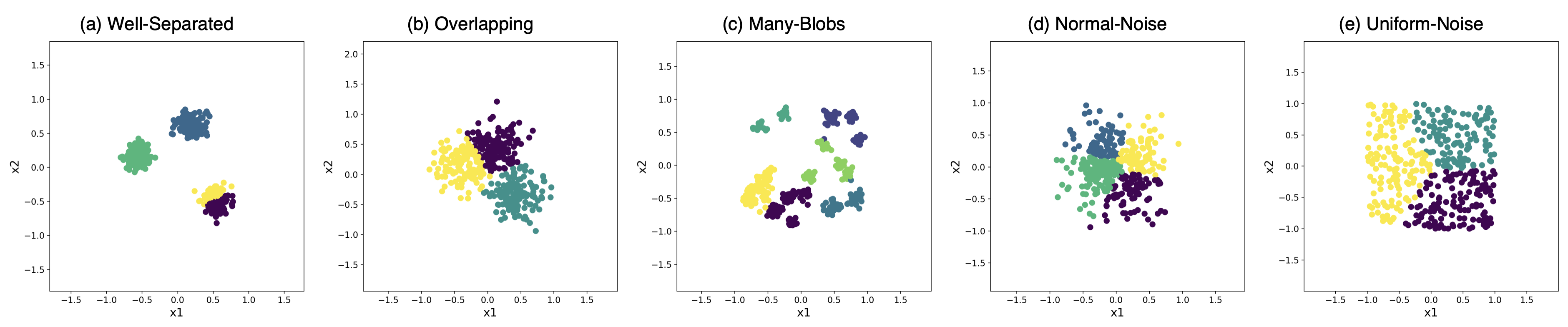}
    \caption{Examples of the five synthetic clustering scenarios used throughout this work: (a) Well-Separated, (b) Overlapping, (c) Many-Blobs, (d) Normal-Noise, and (e) Uniform-Noise.}
    \label{fig:clustering_scenarios}
\end{figure*}

For our dataset, we need a way to test several different common clustering scenarios in a systematic way. To do this, we borrow from Schubert's work \cite{Schubert_2023} and construct five controlled synthetic clustering scenarios that vary in how easily clusters can be assigned. These include Well-Separated, Overlapping, Many-Blobs, Uniform-Noise, and Normal-Noise. We include visual examples of these scenarios in Fig.~\ref{fig:clustering_scenarios}. More specifically, Well-Separated and Overlapping represent clear clusterable structure, differing in how much these groups overlap. Many-Blobs introduces a larger number of smaller groups. Uniform-Noise and Normal-Noise represent increasingly ambiguous cases with no obvious way of assigning clusters. For all experiments, we use Qwen3.5-27B as our VLM \cite{qwen3.5} with thinking disabled and a temperature of $0$.

\subsection{VLM Clustering Evaluation}

Before building the full agentic framework, we first want to test whether our VLM is capable of reliably judging clustering quality from visual inputs. We do this through two exploratory experiments, each one acting as a progressive stepping stone toward the final agentic task.

\begin{figure*}[t]
    \centering

    \begin{subfigure}[t]{0.49\textwidth}
        \centering
        \includegraphics[width=\linewidth]{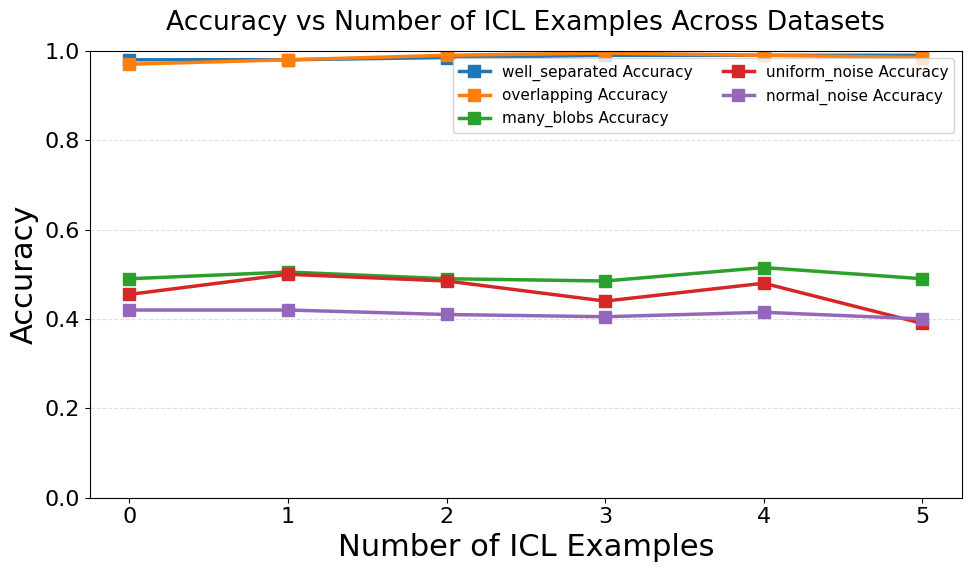}
        \caption{Pairwise clustering comparison accuracy across the five synthetic clustering scenarios as the number of ICL examples increases.}
        \label{fig:pairwise_accuracy}
    \end{subfigure}
    \hfill
    \begin{subfigure}[t]{0.49\textwidth}
        \centering
        \includegraphics[width=\linewidth]{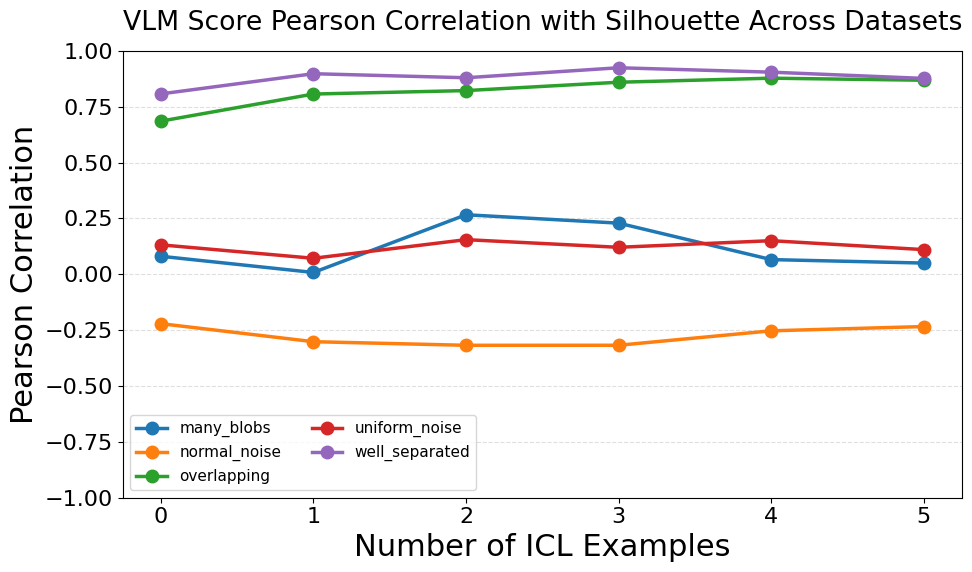}
        \caption{Pearson correlation between VLM clustering scores and the silhouette coefficient across the same clustering scenarios.}
        \label{fig:score_correlation}
    \end{subfigure}

    \caption{Exploratory evaluation of VLM clustering judgment across different numbers of visual in-context examples.}
    \label{fig:exploratory_results}
\end{figure*}

The first of these experiments is a pairwise comparison. Here, the VLM is shown two clustering results side-by-side and is asked which of the two results represents better clustering. We do this for all five of our synthetic scenarios while also varying the number of ICL demonstrations from $0$ to $5$, in order to see how many examples the model needs to be provided with to make an accurate judgment. We then evaluate the model across a total of $1000$ pairwise comparisons, $200$ from each scenario. We determine the "correct" clustering choice by measuring the silhouette coefficient across the two plots and choosing the plot that results in a higher score in order to get a sense of how well the VLM compares to a known metric. As shown in Fig.~\ref{fig:pairwise_accuracy}, the VLM performs almost perfectly on the Well-Separated and Overlapping datasets, while performing considerably worse on Many-Blobs, Uniform-Noise, and Normal-Noise. The results also remain relatively consistent as the number of ICL examples increases. These results suggest that the VLM is able to determine better clustering more easily when there is a clearer separation, similar to that of a person. As the datasets become more ambiguous, we notice that the VLM struggles a lot more, which is to be expected.

The second experiment, being a scoring test, is the more important of these two preliminary studies. The VLM must eventually be able to judge whether a current clustering is good enough to stop iterating by scoring it above a certain threshold. To test this, instead of comparing two plots, we instead show the VLM a single clustering result and ask it to assign a score from $1$ to $10$, with higher scores representing better clustering. We then compare the VLM's scores against the silhouette coefficient using Pearson correlation, again varying the number of ICL demonstrations from $0$ to $5$. We score a total of $1000$ clustering results, $200$ from each scenario. Fig.~\ref{fig:score_correlation}, the VLM's scores have a strong positive correlation with the silhouette coefficient for Well-Separated and Overlapping data, which is surprising since the VLM isn't directly calculating the silhouette coefficient. In comparison, Many-Blobs and Uniform-Noise remain close to zero correlation, while Normal-Noise produces a negative correlation. These results further show that the VLM can make meaningful visual judgments when the cluster structure is clearly visible, while its judgments become less reliable as the structure becomes more ambiguous, similar to that of a person.  

Together, these two experiments give us motivation to move beyond simply evaluating clustering results and test whether the VLM's visual judgment can instead be used as feedback to actively improve an existing clustering algorithm through an agentic loop.

%% file: sec/methods.tex
\section{Methods}

\begin{figure}[t]
    \centering
    \includegraphics[width=0.78\columnwidth]{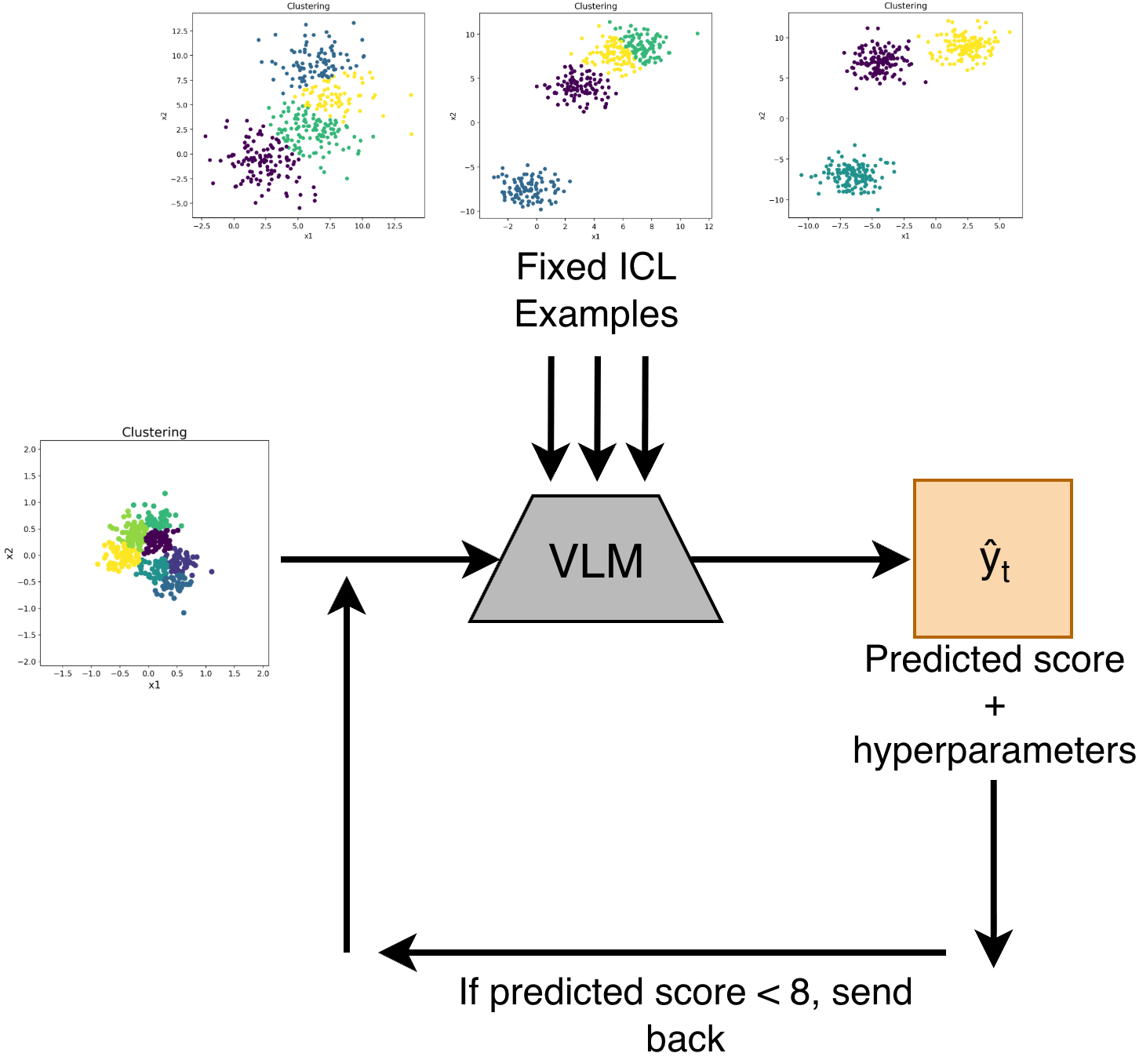}
    \caption{Overview of our agentic clustering framework}
    \label{fig:agentic_framework}
\end{figure}

Having tested the capabilities of our VLM, we move to establishing the core framework of our research--the agentic loop. The first stage of this loop is the introduction of the given clustering in visual form to the overall pipeline. Afterward, the VLM is fed $3$ fixed in-context examples spanning from `worst' to `best' possible clustering, which it then uses to judge the clustering iteration in question. We use $3$ ICL examples because this setting produced the strongest overall Pearson correlation between the VLM scores and silhouette coefficient in our exploratory scoring experiment. This judgment results in the VLM offering a score of how well the given data is clustered (ranging from $1$ to $10$, worst to best), as well as hyperparameter refinements for potentially better k-Means clustering. We visualize the full agentic loop in Fig.~\ref{fig:agentic_framework}.

In our framework, the VLM is allowed to modify the number of clusters $k$ and the k-Means initialization method, choosing between k-means++ and random. For Well-Separated, Overlapping, Uniform-Noise, and Normal-Noise, $k$ can range from $2$ to $10$, while for Many-Blobs we allow values from $2$ to $30$. The VLM score is then compared to the exit condition of the loop, which terminates if the VLM provides a score greater than or equal to $8$. If the VLM reaches this goal, then it exits the loop and outputs the final clustering. If the VLM score is less than the target of $8$, k-Means is rerun using the newly proposed hyperparameters, producing a new clustering visualization that is passed back to the first stage. The loop continues until the sufficient score is met or a predefined maximum of $10$ iterations is reached. To prevent the loop from repeatedly making the same decisions, we also terminate early if the VLM's scores stabilize or if it proposes the same hyperparameters for $3$ consecutive iterations.

%% file: sec/experiments.tex
\section{Experiments}

To evaluate our proposed agentic framework, we test it across the same five synthetic clustering scenarios used in our exploratory study. In this task we use $20$ datasets from each scenario, giving us a total of $100$ clustering tasks. We compare our VLM loop approach against two very common hyperparameter optimization baselines being grid search and Bayesian optimization. Each method searches over the same k-means hyperparameters that our VLM has access to, described in the previous section. Grid search exhaustively evaluates every possible combination, while Bayesian optimization is allowed up to $10$ evaluations. We compare the final clustering performance using the silhouette coefficient for each of these optimization approaches, giving us a way to quantitatively compare our proposed method against these well-established baselines. 

\begin{figure}[t]
    \centering
    \includegraphics[width=1.0\columnwidth]{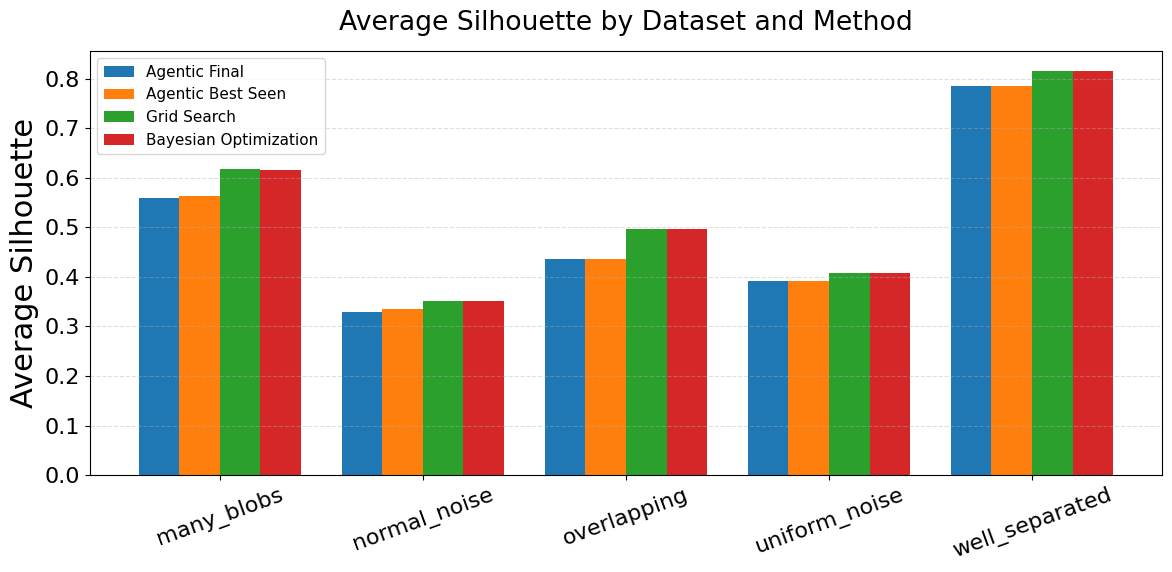}
    \caption{Average silhouette score achieved by the final agentic output, the best configuration seen during the agentic trajectory, Grid Search, and Bayesian Optimization across each synthetic clustering scenario.}
    \label{fig:agentic_silhouette}
\end{figure}

As shown in Fig.~\ref{fig:agentic_silhouette}, our agentic framework produces average silhouette scores that are relatively close to those given through Grid Search and Bayesian Optimization across all of our clustering settings. We show Agentic Final, which represents the clustering when the loop is finished. We also show Agentic Best Seen, which is the highest-silhoette encountered during the agentic loop. Since the VLM could potentially regress and choose hyperparameters that may result in worse clustering, we track best-seen to see how often this happens, and how different it is than the final clustering score. It is important to note that Agentic Best Seen is only used for analysis; the VLM is never given the silhouette coefficient, forcing it to make decisions based on the visuals alone. We can see that Agentic Best Seen and Agentic Final remain extremely close, indicating that once the VLM believes it found the best answer, it stays at that answer.

While grid search and Bayesian optimization typically result in higher silhouette scores, our agentic loop is still able to find hyperparameters that remain relatively close using only the visual of the clustering result along with the ICL examples. This is especially interesting because the baseline methods search for results that directly maximize the silhoette coefficient, while the VLM is only based on visuals. This shows that visual feedback is enough to guide a VLM to find a better solution without directly optimizing one metric.

\begin{figure}[t]
    \centering
    \includegraphics[width=1.0\columnwidth]{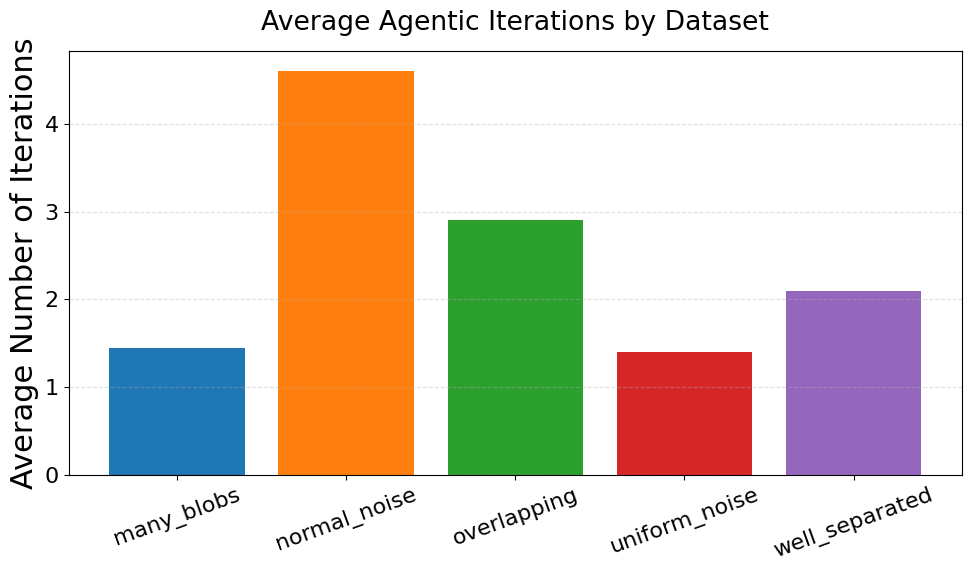}
    \caption{Average number of iterations required by the agentic framework before termination across each synthetic clustering scenario.}
    \label{fig:agentic_iterations}
\end{figure}

Next, we check the number of iterations required by our agentic framework compared to our baselines. In Fig.~\ref{fig:agentic_iterations}, the agent stops in fewer than $5$ iterations on average for every clustering scenario. Uniform-Noise and Many-Blobs require the fewest iterations, while Normal-Noise requires the most, averaging roughly $4.5$ iterations. This result is also consistent with our exploratory experiments, where Normal-Noise was one of the more difficult scenarios for the VLM to evaluate reliably. For comparison, grid search typically requires $18$ evaluations for Well-Separated, Overlapping, Uniform-Noise, and Normal-Noise, and $58$ for Many-Blobs. Bayesian optimization, on the other hand is allowed up to $10$ evaluations. These results demonstrate that the VLM is able to find a somewhat optimal solution with a much smaller amount of iterations. This could particularly be impactful with more hyperparameters to chose from, where in those cases, grid search or Bayesian optimization could be extremely computationally expensive. In those scenarios, this agentic loop may be able to find a relatively optimal solution in fewer iterations without needing to calculate any metrics, which could also be potentially expensive with more points of data.

%% file: sec/conclusion.tex
\section{Conclusion}

The results from our main agentic framework, alongside our two exploratory studies, have proven the capability of an agentic, in-context-driven VLM to improve current clustering algorithm implementations. Our results show that the agentic framework produces clustering configurations with silhouette scores relatively close to those found through established hyperparameter tuning methods such as Grid Search and Bayesian Optimization. Additionally, our framework required fewer than 5 total iterations on average across all our clustering scenarios, far lower than our established baseline methods.

Overall, we view these results as a preliminary proof of concept for using VLMs to evaluate and tune clustering algorithms visually. In this work, we focus only on k-means clustering, while a direct next step would be allowing the VLM to choose between and simultaneously tune different clustering algorithms, such as DBSCAN, HDBSCAN, or spectral clustering. Rather than having a researcher manually inspect many possible combinations of clustering methods, hyperparameters, and plots, the VLM could autonomously explore these choices and suggest the clustering algorithm/procedure that best matches the visual structure. More broadly, this work is an initial step toward using multimodal agents not only to tune clustering algorithms, but potentially to help design new clustering methods. Importantly, all of these algorithms could remain explicit and inspectable, allowing researchers to directly observe decisions and algorithmic choices even if the VLM guiding those decisions is a black box.

%% file: sec/acknowledgments.tex
\section*{Acknowledgments}
Research was supported in part by a UCR Senate Committee on Research grant, the National Science Foundation under grant no. No. 2524228 and CREST Center for Multidisciplinary Research Excellence in CyberPhysical Infrastructure Systems (MECIS) grant no. 2112650 and by the Army Research Office and was accomplished under Grant Number W911NF-24-1-0397. The views and conclusions contained in this document are those of the authors and should not be interpreted as representing the official policies, either expressed or implied, of the Army Research Office or the U.S. Government. The U.S. Government is authorized to reproduce and distribute reprints for Government purposes notwithstanding any copyright notation herein.